\documentclass[letterpaper, 10 pt, conference]{styles/ieeeconf}  

\IEEEoverridecommandlockouts                              

\usepackage{graphicx} 
\usepackage{xcolor}
\usepackage{amsmath} 
\usepackage{amssymb}  
\usepackage{algorithm}
\usepackage{algpseudocode}
\usepackage{hyperref}
\usepackage{tikz}
\usepackage{booktabs}
\usepackage{makecell}
\usetikzlibrary{shapes, arrows.meta, positioning, calc, fit, backgrounds, decorations.markings}
\definecolor{netblue}{RGB}{220, 240, 255}
\definecolor{lossred}{RGB}{255, 220, 220}
\definecolor{inputgreen}{RGB}{220, 255, 220}
\definecolor{dataorange}{RGB}{255, 240, 200}

\newcommand\copyrighttext{%
  \footnotesize \textcopyright 2026 IEEE. Personal use of this material is permitted.
  Permission from IEEE must be obtained for all other uses, in any current or future
  media, including reprinting/republishing this material for advertising or promotional
  purposes, creating new collective works, for resale or redistribution to servers or
  lists, or reuse of any copyrighted component of this work in other works.}
\newcommand\copyrightnotice{%
\begin{tikzpicture}[remember picture,overlay]
\node[anchor=south,yshift=10pt] at (current page.south) 
  {\fbox{\parbox{\dimexpr\textwidth-\fboxsep-\fboxrule\relax}{\copyrighttext}}};
\end{tikzpicture}%
}

\title{\LARGE \bf
End-to-End Safe Social Navigation via Multi-Task Reinforcement Learning and Probabilistic Perception}

\author{Tommaso Van Der Meer$^{1}$, Andrea Garulli$^{1}$, Antonio Giannitrapani$^{1}$, Renato Quartullo$^{2}$, Alberto Vaglio$^{1}$\\ and Alexandre Alahi$^{3}$%
\thanks{$^{1}$ Dipartimento di Ingegneria dell'Informazione e Scienze Matematiche, Università di Siena, Siena, Italy. E-mail: \{tommaso.vandermeer, andrea.garulli, antonio.giannitrapani, alberto.vaglio\}@unisi.it.}%
\thanks{$^{2}$ Uninettuno University, Rome, Italy. E-mail: 
renato.quartullo@uninettunouniversity.net.}%
\thanks{$^{3}$ Visual Intelligence for Transportation (VITA) Lab, École Polytechnique Fédérale de Lausanne (EPFL),
Lausanne, Switzerland. E-mail: alexandre.alahi@epfl.ch}%
}

\begin{document}

\maketitle
\copyrightnotice
\thispagestyle{empty}
\pagestyle{empty}

\begin{abstract}

Autonomous social navigation requires balancing efficiency, physical safety, and social compliance. Reinforcement Learning (RL) methods provide a viable and effective solution but often rely on unrealistic assumptions, such as the knowledge of humans' position and velocity. In this paper, we introduce JESSI (JAX-based E2E Safe Social Interpretable navigation), a lightweight end-to-end RL framework that maps raw LiDAR scans directly to kinematically feasible control commands. JESSI enhances safety via Dirichlet-parameterized continuous action spaces and deterministic bounding, while an integrated attention-based perception module extracts probabilistic human states for interpretable, socially aware decision-making. Through extensive simulations and real-world deployment on a differential-drive robot, we demonstrate that jointly optimizing the RL policy with a supervised perception signal in a multi-task paradigm enhances social behavior. Ultimately, JESSI is able to balance high navigation success rates and superior social behaviors compared to state-of-the-art baselines.

\end{abstract}


\section{INTRODUCTION}

The deployment of autonomous mobile robots in unstructured, human-populated environments has seen tremendous progress in recent years. Navigating effectively in these scenarios is a challenging task. It requires a delicate balance between navigation efficiency, strict physical safety (e.g., avoiding collisions with walls and pedestrians), and adherence to unwritten social norms (e.g., maintaining comfortable distances) \cite{kruse2013human, mavrogiannis2023core}. While classical model-based planning algorithms have long been the standard, they often struggle to capture the complex, interactive nature of crowd dynamics \cite{trautman2010unfreezing}. Consequently, Reinforcement Learning (RL) has emerged as a promising paradigm, allowing robots to learn sophisticated navigation policies directly from the interaction with the environment.

Despite achieving impressive results in simulated environments, state-of-the-art RL and model-based social navigation approaches often rely on strong, and sometimes unrealistic, assumptions that inhibit their real-world applicability:

\emph{Perfect state observability}: many top-performing algorithms assume the availability of the exact state of all surrounding humans (e.g., precise positions and velocities) at any given time \cite{CADRL, SARL, li2019sarl, trevisan2025dynamic, van2025safe}. In reality, onboard perception is limited by sensor noise, limited fields of view, and severe occlusions.

\emph{Omission of static obstacles}: several foundational social navigation policies focus exclusively on human-robot interaction, omitting static obstacles from their formulation or training environments \cite{CADRL, SARL}. This renders them inapplicable in typical, cluttered indoor spaces.

\emph{Humans as generic obstacles}: conversely, navigation methods designed to operate directly from raw sensor data or classical dynamics often treat humans merely as moving geometric shapes \cite{fox2002dynamic, williams2017model, pfeiffer2017perception, dugas2021navrep}. By lacking an explicit semantic understanding of human presence, these methods produce navigation policies that are highly optimized for collision avoidance but exhibit poor social behavior.

\emph{Kinematic infeasibility}: many continuous-action RL algorithms generate commands that do not intrinsically respect the kinematic of the robot, relying instead on artificial a-posteriori velocity clipping which can destabilize learning and execution \cite{fujita2018clipped, hwangbo2019learning}.

\begin{figure}[t]
    \centering
    \includegraphics[width=.95\columnwidth]{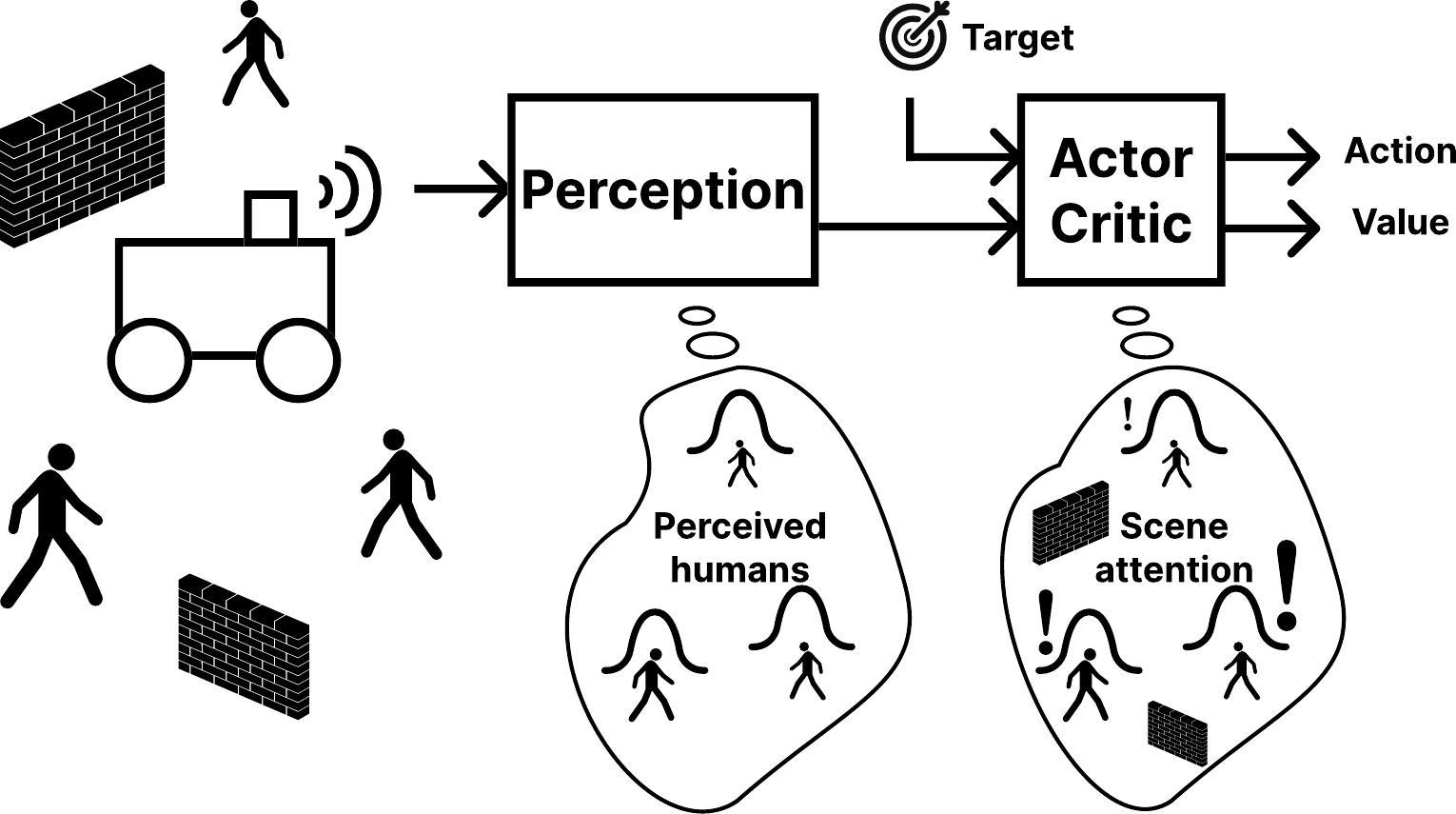} 
    \caption{Overview of the proposed method. JESSI operates directly from raw LiDAR scans and leverages a perception module to extract probabilistic human states (human-centered bivariate gaussians) and plan socially-aware and kinematically feasible control commands through a scene attention mechanism and Dirichlet distributions.}
    \label{fig:teaser}\vspace{-4mm}
\end{figure}

In the broader context of autonomous driving, recent breakthroughs have demonstrated that purely reactive End-to-End (E2E) systems, which directly map sensor inputs to control commands, often suffer from poor sample efficiency and lack of interpretability \cite{chen2024end}. To overcome this, the community adopted multi-task learning paradigms, demonstrating that jointly optimizing auxiliary perception tasks alongside the primary planning objective significantly enhances the overall navigation performance \cite{prakash2021multi,hu2023planning}.

Inspired by these advancements, we introduce \textbf{JESSI} (\textit{JAX-based E2E Safe Social Interpretable navigation}). JESSI is a novel reinforcement learning framework designed to bring the benefits of multi-task E2E learning into the realm of safe social navigation. An overview of the proposed method is shown in Fig. \ref{fig:teaser}. Its core design principles are:

\emph{End-to-end}: breaking away from the assumption of perfect state observability, our architecture cascades a perception network and an actor-critic policy to compute control actions directly from raw LiDAR scans. Bypassing the need for an external oracle to provide humans' positions and velocities, the system natively handles real-world sensor noise, limited fields of view, occlusions, and also considers static obstacles for motion planning. Moreover, the network is agnostic to the specific LiDAR setting implemented.

\emph{Safety}: to ensure physical safety and strictly respect the kinematic constraints of a differential-drive robot, JESSI maps the continuous action space using Dirichlet distributions \cite{ma2024constrained}. This is coupled with a deterministic bounding mechanism \cite{van2025safe} that prevents actions leading to collisions with static obstacles.

\emph{Social compliance}: unlike standard raw-sensor end-to-end methods, JESSI explicitly models human presence. The perception module extracts spatial and velocity estimates of surrounding pedestrians from the LiDAR stack. By feeding this structured information into the policy's scene attention layer, the robot is intrinsically encouraged to learn socially compliant behaviors.

\emph{Interpretability}: JESSI does not implement a purely black-box network but uses an intermediate representation layer. The perception module explicitly outputs confidence scores and bivariate Gaussian distributions representing the states of visible humans. This not only grounds the RL policy on a physically meaningful latent space but also provides useful insight into the robot decisions. However, differently from pure modular pipelines, the overall architecture maintains full differentiability, allowing for different training paradigms to be employed.

\emph{Efficiency}: despite integrating a dedicated perception module, JESSI features a deliberately lightweight neural architecture. Implemented using the JAX library \cite{jax2018github}, it leverages hardware-accelerated vectorization and just-in-time (JIT) compilation. This combination of compact architecture and JAX compilation enables efficient inference, making the policy suitable for real-time deployment on embedded systems.

The primary contribution of this work is a hybrid navigation framework that combines the strengths of end-to-end and modular architectures. This structure provides the explicit interpretability of modular pipelines while preserving total differentiability for joint optimization. In addition, the adoption of Dirichlet distributions, together with a deterministic action space bounding algorithm introduces a rigorous safety layer that ensures kinematic feasibility and collision avoidance at the control level.

The paper is organized as follows: Sec.\ref{sec:method} presents the considered robot kinematic model and the proposed method, Sec. \ref{sec:numerical_results} showcases the simulated environment employed and the simulation experiments performed against various baselines. Sec. \ref{sec:real_world} introduces the experiments carried out in the real world on two different robotic platforms, and Sec. \ref{sec:conclusions} draws some conclusions giving useful insights on the proposed algorithm.

\section{METHOD}\label{sec:method}

\begin{figure}[t]
    \centering
    \includegraphics[width=\columnwidth]{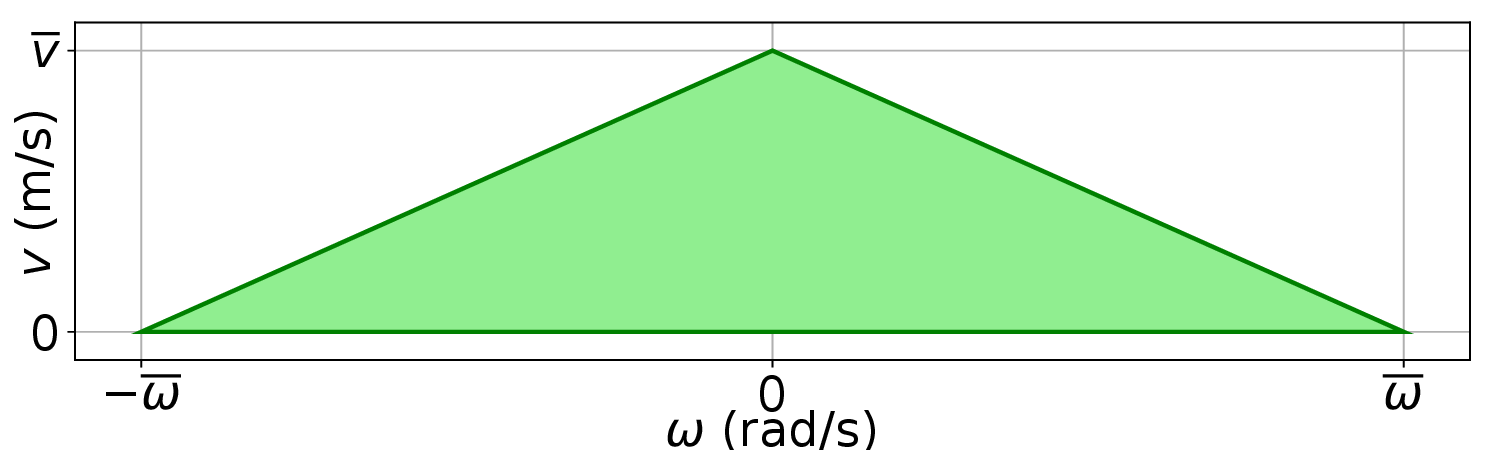} 
    \caption{Triangular action space of the differential drive with linear velocity $v \in [0, \bar{v}]$ and angular velocity $\omega \in [-\bar{\omega}, \bar{\omega}]$.}
    \label{fig:action_space}\vspace{-5mm}
\end{figure}

\begin{figure*}[thpb]
\centering
\resizebox{\textwidth}{!}{
    \begin{tikzpicture}[
        >=stealth,
        font=\sffamily\small,
        input/.style={rectangle, draw=green!60!black, fill=green!10, rounded corners, thick, align=center, text width=2.4cm, minimum height=1.5cm},
        layerP/.style={rectangle, draw=blue!60, fill=blue!10, thick, align=center, text width=2.4cm, minimum height=1.5cm},
        layerAC/.style={rectangle, draw=red!60, fill=red!10, thick, align=center, text width=2.4cm, minimum height=1.5cm},
        data/.style={rectangle, draw=orange!80!black, fill=orange!10, thick, align=center, text width=2.4cm, minimum height=1.5cm},
        output/.style={rectangle, draw=purple!80, fill=purple!10, rounded corners, thick, align=center, text width=2.0cm, minimum height=1.4cm},
        concat/.style={circle, draw=black, thick, fill=gray!20, inner sep=0pt, minimum size=0.6cm, font=\large\bfseries},
        layerP_exp/.style={trapezium, shape border rotate=90, trapezium angle=80, draw=blue!60, fill=blue!10, thick, align=center, text width=2.2cm, minimum height=2.6cm, inner sep=2pt},
        layerP_dec/.style={trapezium, shape border rotate=270, trapezium angle=80, draw=blue!60, fill=blue!10, thick, align=center, text width=2.2cm, minimum height=2.6cm, inner sep=2pt},
        layerAC_dec/.style={trapezium, shape border rotate=270, trapezium angle=80, draw=red!60, fill=red!10, thick, align=center, text width=2.2cm, minimum height=2.6cm, inner sep=2pt},
        arrow/.style={->, thick, draw=black!80},
        dashed_arrow/.style={->, thick, dashed, draw=black!80},
        arr_txt/.style={midway, above, font=\sffamily\footnotesize, text=black!80, align=center}
    ]
    \def\xdist{3.8}
    \node[input]      (lidar)      at (0, 0) {\textbf{LiDAR stack}};
    \node[layerP_exp] (mlp)        at (\xdist*1, 0) {\textbf{Shared MLP}\\Token embed.};
    \node[layerP]     (ang_attn)   at (\xdist*2, 0) {\textbf{Angular local}\\ \textbf{cross-attn.}\\(Parallel per scan)};
    \node[layerP]     (st_attn)    at (\xdist*3, 0) {\textbf{Spatio-temp.}\\ \textbf{Self-Attn}\\Flatten time};
    \node[layerP_dec] (cross_attn) at (\xdist*4, 0) {\textbf{Cross-attn.}\\$N_k$ Queries};
    \node[layerP]     (self_attn)  at (\xdist*5, 0) {\textbf{Self-attn.}};
    \node[layerP_dec] (fc_perc)    at (\xdist*6, 0) {\textbf{Shared MLP}\\Dim reduction};
    \node[data]       (humans)     at (\xdist*7, 0) {\textbf{Human states}};
    
    \def\yAC{-4.8}
    \node[input]       (proprio)     at (0, \yAC) {\textbf{Proprioception}\\State \& kin.};
    \node[data]        (latest_scan) at (\xdist*2, \yAC) {\textbf{Latest scan}};
    \node[layerAC_dec] (fc_lidar)    at (\xdist*3, \yAC) {\textbf{Shared MLP}\\Dim. reduction};
    \node[layerAC]     (scene_attn)  at (\xdist*4, \yAC) {\textbf{Scene}\\ \textbf{self-attn.}};
    \node[layerAC_dec]     (pooling)     at (\xdist*5, \yAC) {\textbf{Weighted}\\ \textbf{pooling}};
    \node[concat]      (concat_node) at (\xdist*6, \yAC) {$\oplus$};
    \node[layerAC_dec] (actor)  at (\xdist*7, \yAC + 0.9) {\textbf{Actor net.}\\MLP};
    \node[layerAC_dec] (critic) at (\xdist*7, \yAC - 0.9) {\textbf{Critic net.}\\MLP};
    \node[output, right=0.6cm of actor]  (action) {\textbf{Action}\\$\boldsymbol{\delta}$ (Dirichlet)};
    \node[output, right=0.6cm of critic] (value)  {\textbf{Value}\\$V$};
    
    \draw[arrow] (lidar) -- (mlp) node[arr_txt] {$(N_t, N_b,$\\$ 7)$};
    \draw[arrow] (mlp) -- (ang_attn) node[arr_txt] {$(N_t, N_b,$\\$N_e)$};
    \draw[arrow] (ang_attn) -- (st_attn) node[arr_txt] {$(N_t, N_s,$\\$N_e)$};
    \draw[arrow] (st_attn) -- (cross_attn) node[arr_txt] {$(N_t \times N_s,$\\$N_e)$};
    \draw[arrow] (cross_attn) -- (self_attn) node[arr_txt] {$(N_k, N_e)$};
    \draw[arrow] (self_attn) -- (fc_perc) node[arr_txt] {$(N_k, N_e)$};
    \draw[arrow] (fc_perc) -- (humans) node[arr_txt] {$(N_k, 11)$};
    
    \draw[arrow] (ang_attn) -- (latest_scan) node[midway, right, font=\footnotesize] {Extract latest scan};
    \draw[arrow] (humans.south) -- ++(0, -1.2) -| (scene_attn.north) node[near start, above, font=\footnotesize] {$(N_k, 11)$ Features};
    \draw[dashed_arrow] (humans.south) -- ++(0, -2.1) -| (pooling.north) node[near start, above, font=\footnotesize] {Confidence scores $\{s_k\}$};
    \draw[arrow] (latest_scan) -- (fc_lidar) node[arr_txt] {$(N_s, N_e)$};
    \draw[arrow] (fc_lidar) -- (scene_attn) node[arr_txt] {$(N_s)$};
    \draw[arrow] (scene_attn) -- (pooling);
    \draw[arrow] (pooling) -- (concat_node);
    \draw[arrow] (fc_lidar.north) -- ++(0, 0.8) -| (concat_node.north) node[near start, above, font=\footnotesize] {Obstacle Features $(N_s)$};
    \draw[arrow] (proprio.south) -- ++(0, -0.8) coordinate (prop_bus) -| (scene_attn.south);
    \draw[arrow] (prop_bus) -| (concat_node.south) node[near start, above, font=\footnotesize] {Robot State};
    \draw[arrow] (concat_node.east) -- ++(0.3, 0) |- (actor.west);
    \draw[arrow] (concat_node.east) -- ++(0.3, 0) |- (critic.west);
    \draw[arrow] (actor) -- (action);
    \draw[arrow] (critic) -- (value);
    
    \begin{scope}[on background layer]
        \coordinate (box_left)  at ($(lidar.west) + (-0.4, 0)$);
        \coordinate (box_right) at ($(action.east) + (0.4, 0)$);
        
        \node[rectangle, draw=blue!80, fill=blue!3, dashed, thick, rounded corners, 
              fit=(box_left |- lidar.north) (box_right |- humans.south) (ang_attn.north), 
              inner sep=12pt, label={[font=\Large\bfseries, text=blue!80]above:Perception Module}] (box_perc) {};
        
        \node[rectangle, draw=red!80, fill=red!3, dashed, thick, rounded corners, 
              fit=(box_left |- actor.north) (box_right |- prop_bus) (fc_lidar.north), 
              inner sep=12pt, label={[font=\Large\bfseries, text=red!80]below:Actor-Critic Module}] (box_ac) {};
    \end{scope}
    \end{tikzpicture}
}
\caption{Complete architecture of the JESSI policy. The top part in blue represents the perception module while the bottom part in red depicts the actor-critic module. The inputs of the overall network are depicted in green blocks. In order, the dimensionalities $N_t$, $N_b$, $N_e$, $N_s$, $N_k$ represent, the LiDAR stack dimension, the number of LiDAR beams for each stack, the embedding size, the number of angular sectors of the cross-attention layer, and the number of objects predicted by the perception module.}
\label{fig:jessi_architecture}\vspace{-5mm}
\end{figure*}
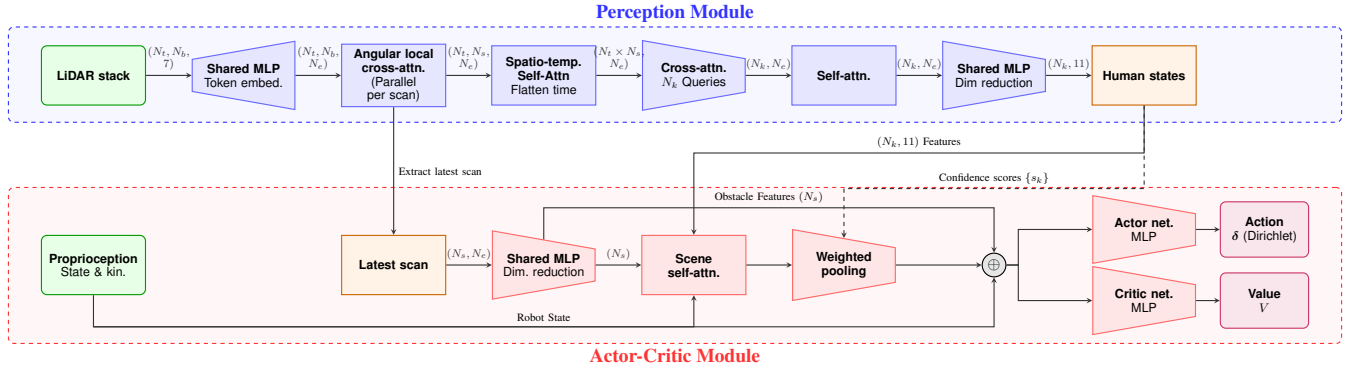

The method relies on a continuous action space parameterization that inherently respects the robot's kinematic constraints, an attention-based perception module capable of extracting probabilistic representations of the state of surrounding humans from raw sensor data, and a multi-stage training pipeline to ensure sample efficiency and stable convergence.

\subsection{Robot kinematics and action space parameterization}
We consider a differential-drive mobile robot modeled as a disk of radius $r$. Let $p=[x,y]^T$ and $\theta$ denote the position and orientation of the robot in a global reference frame, respectively. The discrete-time kinematic model of the robot is given by \cite{thrun2002probabilistic}:\vspace{-2mm}
\begin{align}
    x^{+} &= x + \frac{v}{\omega}(\sin(\theta+\omega\Delta t)-\sin(\theta)) \\
    y^{+} &= y + \frac{v}{\omega}(\cos(\theta)-\cos(\theta+\omega\Delta t)) \\
    \theta^{+} &= \theta + \omega\Delta t
\end{align}
where $v$ and $\omega$ are the linear and angular speeds representing the action $a=[v, \omega]^T$, $\Delta t$ is the sampling time, and the superscript $+$ denotes the quantities at the next time step $t+\Delta t$. Due to the physical limitations of the wheel actuators, the rotational speeds of the left and right wheels are bounded, and the robot is forced to execute forward motion only ($v \ge 0$). These constraints naturally define a triangular feasible continuous action space in the $(v, \omega)$ plane (see Fig. \ref{fig:action_space}).

To map the outputs of the neural network directly into the feasible space without incurring on constraint violations, Dirichlet distributions over the vertices of the triangle are used to parameterize the policy action space, as introduced in \cite{ma2024constrained}. Dirichlet distributions $\text{Dir}(\boldsymbol{\delta})$ are parameterized by the parameter vector $\boldsymbol{\delta}$ and can be used to model any convex action space. Using this paradigm, actions are computed as $\mathbf{a} = \mathbf{V w}$, where $\mathbf{V}$ is the matrix of the vertices of the considered action space and $\mathbf{w} \sim \text{Dir}(\boldsymbol{\delta})$ is a sample of the Dirichlet distribution.
Furthermore, to promote collision avoidance, the lightweight bounding algorithm proposed in \cite{van2025safe} is employed. In this work, the vertices of the triangular action space are scaled using the latest LiDAR scan, to prevent intersections between the next robot configuration and the measured point cloud.

\subsection{Neural network architecture}
The core of the proposed approach is an integrated neural network including a perception module and an actor-critic policy, designed to process raw sensory data and output safe navigation commands. The overall architecture is detailed in Fig. \ref{fig:jessi_architecture}.

\subsubsection{Perception Module}
The perception module processes a stack of the latest $N_t$ LiDAR scans, each one consisting of $N_b$ beams aligned to the last robot-centric frame to predict the dynamic state of nearby pedestrians. Specifically, the input is re-parameterized to include the following information $(d^s_b, \text{hit}^s_b,x^s_b,y^s_b,\sin{\theta^s_b},\cos{\theta^s_b},\Delta t^s)$ for each beam $b$ of each scan $s$. In order, these elements represent the detected distance normalized by the maximum distance range ($d^s_b \in [0,1]$), a boolean indicating whether the measurement is within the feasible range, the Cartesian coordinates of the hit-point, the sine and cosine of the beam angle in the latest robot-centric frame, and the difference between the measurement time of the latest scan and the time when the reading was taken. The human perception process is tackled by drawing inspiration from recent advancements in transformer-based object detection \cite{carion2020end}. First, each beam is projected into a higher-dimensional space through a shared multi-layer perceptron that produces a beam token embedding of dimension $N_e$. To ensure the architecture is fully agnostic to the specific hardware configuration of the LiDAR (e.g., resolution and field of view), an angular local cross-attention mechanism is employed \cite{vaswani2017attention, liu2021swin}. The incoming $N_t \times N_b$ beam token embeddings act as keys and values, while a set of learnable latent $N_s$ vectors, evenly distributed in the angular space and encoding robot-fixed angular sectors, act as object queries. To enforce spatial consistency, an angular mask restricts the attention weights, ensuring that latent queries only attend to LiDAR beams within their specific angular sector. Each scan in the stack is processed in parallel through the same angular local cross-attention layer. Subsequently, the stack (or time) dimension is flattened and the $N_t \times N_s$ features are passed through a spatio-temporal self-attention layer that aggregates the spatial features across the temporal stack, making the network invariant to the specific length of the observation history. The output of this layer is then passed to another cross-attention layer with learned queries to lower the dimension from $N_s \times N_t$ to the desired number of output components $N_k$. Later, a self-attention layer is employed to prevent multiple output components from matching the same human in the environment and finally a shared fully connected network lowers the dimension of each component embedding to the final size. Regarding the output formulation, we adopt the ``objects as points'' philosophy \cite{zhou2019objects}. Each pedestrian is modeled as a probabilistic center point, allowing the network to directly output a set of full bivariate Gaussian distributions representing the estimated positions, velocities, and spatial uncertainties of the humans in the scene. The network outputs $N_k$ components including a confidence score $s_k$ encoding the probability of that component being a real human, a bivariate Gaussian encoding the position $(\mu^p_{k,x}, \mu^p_{k,y}, \sigma^p_{k,xx}, \sigma^p_{k,yy}, \rho^p_k)$, a bivariate Gaussian encoding the velocity $(\mu^v_{k,x}, \mu^v_{k,y}, \sigma^v_{k,xx}, \sigma^v_{k,yy}, \rho^v_k)$ for each predicted component $k$. This formulation inherently captures the spatial uncertainty caused by LiDAR occlusions and sparsity. The perception solves a detection problem rather than a tracking problem, as the downstream actor-critic module does not use humans' identity but only their probabilistic state (encoding positions and velocities).

\subsubsection{Actor-Critic policy}
The policy leverages an actor-critic architecture. The inputs to the networks are three: the robot's proprioceptive data (the polar coordinates of the goal in the robot frame, robot radius, maximum linear velocity and distance between the wheels), the probabilistic $N_k$ human states generated by the perception module and the latest LiDAR scan of dimension $(N_s, N_e)$ already processed by the angular local cross-attention of the perception module, which encodes the spatial features of the obstacles in the scene. Note that, instead of using just the mean values of the humans' position and velocities (as in \cite{CADRL,SARL,van2025safe}), also their uncertainties are fed into the network (since the perception module returns also the covariance matrices of the estimated humans' position and velocity).
First, the latest LiDAR scan is passed through a shared fully connected network to lower the dimension of each angular sector embedding. Then, a scene self attention mechanism is employed to predict which human to pay more attention to, based on the whole scene context (humans, robot and obstacles).
Later, the incoming features are aggregated via a weighted pooling mechanism, where the weights are directly derived from the $N_k$ confidence scores $s_k$ produced by the perception module. This design forces the policy to heavily rely on highly certain human detections while suppressing noisy predictions (i.e., components with low score).
Finally, the pooled human features are concatenated with the robot state and the latest scan features before being routed to the actor and critic networks. The actor parametrizes the Dirichlet distribution by outputting the concentration parameters $\boldsymbol{\delta}$ to sample the continuous action, whereas the critic estimates the value of the current state.

\subsection{Multi-stage training pipeline}\label{sec:multi_stage_training}
Training a complex, end-to-end architecture directly with RL from sparse rewards is notoriously sample-inefficient and prone to sub-optimal convergence. To overcome this, a three-stage training scheme is adopted.

\textbf{Perception pre-training:} in the first stage, only the perception module is trained in a supervised learning fashion using ground-truth data from many simulated episodes. Since the network predicts an unordered set of $K$ potential humans, the Hungarian algorithm \cite{kuhn1955hungarian} is first applied to establish a bipartite matching between predictions and ground-truth targets. The perception loss $\mathcal{L}_{perc}$ is then computed as a weighted sum:
\begin{equation}\label{eq:perc_loss}
    \mathcal{L}_{perc} = \lambda_{pos} \mathcal{L}_{NLL}^{pos} + \lambda_{vel} \mathcal{L}_{NLL}^{vel} + \lambda_{cls} \mathcal{L}_{BCE}^{cls}
\end{equation}
where $\mathcal{L}_{NLL}^{pos}$ and $\mathcal{L}_{NLL}^{vel}$ are the negative log-likelihoods of the true positions and velocities evaluated under the predicted bivariate Gaussians, and $\mathcal{L}_{BCE}^{cls}$ is the binary cross-entropy loss penalizing the classification of the confidence scores. In this phase, data augmentation techniques such as random rotations and Gaussian noise are applied to the input to improve regularization during training.
    
\textbf{Imitation learning warm-up:} in the second stage, the actor-critic networks undergo pre-training via behavioral cloning. We utilize a trained DIR-SAFE local planner \cite{van2025safe} to generate expert trajectories (but any well-established planner can be used for such purpose). This generates an initial policy able to move towards the goal, effectively contrasting the initial complexity in tracking sparse reward signals. The Imitation Learning (IL) loss $\mathcal{L}_{IL}$ minimizes the Mean Squared Error (MSE) for both the actor and the critic:
    \begin{equation}\label{eq:il_loss}
        \mathcal{L}_{IL} = \mathcal{L}^{A}_{MSE} + \lambda_{V}  \, \mathcal{L}^{V}_{MSE}
    \end{equation}
where $\mathcal{L}^{A}_{MSE}$ is the MSE loss to match the expert action with the mean action computed from the predicted Dirichlet distribution and $\mathcal{L}^{V}_{MSE}$ is the loss to match the predicted state value and the empirical discounted return for each state.
    
\textbf{End-to-end reinforcement learning:} in the final stage, the entire architecture is fine-tuned concurrently. The Proximal Policy Optimization (PPO) algorithm is employed \cite{schulman2017proximal}. To enforce robust representation learning, the auxiliary perception loss is injected into the total objective $\mathcal{L}_{total}$:
\begin{equation}\label{eq:rl_loss}
    \mathcal{L}_{total} = \mathcal{L}_{PPO}^{A} + \lambda_{V} \mathcal{L}_{PPO}^{V} - \lambda_{ent}  \mathcal{L}_{ent} + \lambda_{aux} \mathcal{L}_{perc}
\end{equation}
where $\mathcal{L}_{PPO}^{A}$ is the PPO clipped surrogate objective, $\mathcal{L}_{PPO}^{V}$ is the clipped value function squared error and $\mathcal{L}_{ent}$ is the Dirichlet entropy bonus encouraging exploration. The weighting coefficients $\lambda_{V}$, $\lambda_{aux}$ and $\lambda_{ent}$ serve to scale the respective loss terms. Depending on the specific training paradigm (as detailed in Section \ref{sec:training_paradigms}), the gradients from $\mathcal{L}_{PPO}^{A}$ and $\mathcal{L}_{PPO}^{V}$ may or may not be allowed to flow back through the perception backbone.

\section{NUMERICAL RESULTS}\label{sec:numerical_results}

To evaluate the effectiveness of the proposed approach, an extensive campaign of simulated experiments is conducted. The simulation environment and the scenario design are first described.

\subsection{Simulation environment and scenario design}
Training and evaluating social navigation policies requires an environment capable of generating realistic, diverse, and interactive pedestrian behaviors. To achieve this, a highly parallelized 2D simulator has been developed. 
Rather than relying on simple constant-velocity models for the pedestrians, all simulated humans are governed by the Headed Social Force Model (HSFM) \cite{farina2017walking}, which extends the classic social force model \cite{helbing2000simulating} by explicitly accounting for the pedestrians' heading. This generates fairly realistic trajectories where humans naturally yield, decelerate, and avoid collisions with both static obstacles and each other. To obtain trained policies which take full responsibility on avoiding humans, the robot is not visible to them during training. Conversely, during testing, the robot is visible to humans in order to make the environment more realistic.

To ensure the evaluation is comprehensive and aligned with the current standard practices in the community, the topologies of the training and testing environments are inspired by the taxonomy proposed in \cite{francis2023principles}. 
A rich suite of scenarios has been designed, covering both symmetric and asymmetric interactions, such as perpendicular crossings, parallel traffic, bottlenecks (e.g., doors), and dense crowds. In addition, the simulator randomizes the initial placement of humans and static obstacles, forcing the policy to learn robust generalization rather than memorizing specific spatial layouts. Finally, the LiDAR inputs (captured within a 360° field-of-view with a maximum range of 10 meters) are corrupted using both small additive and multiplicative Gaussian noise and some measurements are randomly overwritten with the maximum distance range, to simulate the presence of outliers. The training environment includes five humans and three static obstacles. The dimensionalities chosen for the architecture of JESSI are: $N_t=5$, $N_b=100$, $N_e=96$, $N_s=60$ and $N_k=10$. However, given the sensor-agnostic architecture of the perception module, the trained network can be operated with any number of LiDAR scans $N_t$ and any number of LiDAR beams per scan $N_b$ (possibly at the cost of reduced performance or increased inference time).

\subsection{Reward function}
The reinforcement learning objective is to maximize the expected cumulative reward. We adopt a reward formulation which encourages progress towards the goal, penalizes collisions, and promotes socially compliant and smooth trajectories. The total reward $R$ is defined as a weighted sum of three components:
\begin{equation}
    R = w_b R_{base} + w_p R_{prog} + w_r R_{rot}
\end{equation}
The base reward $R_{base}$, also employed in \cite{CADRL, SARL, van2025safe}, handles goal reaching, collision penalties, and social comfort. Specifically, $R_{base}$ is defined as:
\begin{equation}
    R_{base} = 
    \begin{cases} 
      -0.5 & \text{if } d^h_{min} \leq 0 \text{ (human collision)}\\
      -0.05 & \text{if } d^o_{min} \leq 0 \text{ (obstacle collision)}\\
      -\frac{(d_{s} - d^h_{min})\Delta t}{2} & \text{if } 0 <  d^h_{min} < d_{s} \\
      1 & \text{if } ||\mathbf{p}^+ - \mathbf{p}^{goal}|| < r \\
      0 & \text{otherwise}
   \end{cases}
\end{equation}
where $d^h_{min}$ is the minimum distance between the robot boundary and all humans, $d^o_{min}$ is the minimum distance between the robot boundary and all obstacles, $d_s$ is the social discomfort distance, $r$ is the robot radius, and $\mathbf{p}^+$ and $\mathbf{p}^{goal}$ are the next robot position and the goal position, respectively. The term $R_{prog}$ rewards the reduction of distance to the goal, while $R_{rot}$ penalizes high angular velocities to promote smoother paths. The reward parameters are chosen as  $w_b = 1$, $w_p = 0.03$, $w_r = 0.0075$, $d_s = 0.2$~m and $\Delta t = 0.25$~s.

\subsection{Training paradigms}\label{sec:training_paradigms}
To investigate how the gradients from the various loss terms influence the representation learning of each module, JESSI end-to-end reinforcement learning stage (see Section \ref{sec:multi_stage_training}) is carried out under three distinct paradigms. In all cases, the policy is optimized using PPO \cite{schulman2017proximal}.

\textbf{JESSI-MT (Multi-Task):} the network is trained end-to-end using a weighted sum of the RL loss and the perception loss, as detailed in Eq. \eqref{eq:rl_loss}. Gradients from both objectives flow all the way back through the entire network. This allows the perception module to learn task-aware representations, potentially focusing more on humans that are critical for the navigation task.
    
\textbf{JESSI-MD (Modular):} the total loss remains a weighted sum of the RL and perception objectives (Eq. \eqref{eq:rl_loss}). However, a stop-gradient operator is placed at the input of the actor-critic networks. Consequently, the RL gradients do not propagate into the perception module. The perception backbone is updated exclusively by the supervised perception loss. This paradigm prevents the RL signal from changing the state estimation but relies on an additional training phase to fine-tune its estimates, during policy training.
    
\textbf{JESSI-PO (Policy-Only):} the perception module is completely frozen after its initial supervised pre-training phase, acting as a fixed feature extractor. The training process relies solely on the RL loss (Eq. \eqref{eq:rl_loss} with $\lambda_{aux} = 0$), which is used to update only the weights of the actor-critic networks. This represents the most computationally lightweight training approach.

The weights associated to each loss term (if present) in \eqref{eq:perc_loss}, \eqref{eq:il_loss} and \eqref{eq:rl_loss} are chosen as: $\lambda_{V} = 0.5$, $\lambda_{aux} = 0.05$, $\lambda_{ent} = 0.0005$, $\lambda_{pos}=2.0$, $\lambda_{vel}=1.0$ and $\lambda_{cls}=1.0$. The pre-trainining of the perception module requires approximately 20 hours, the the imitation learning warm-up on the actor-critic module takes around 10 minutes and the RL fine tuning needs approximately 7 hours. All the learning processes have been carried out on a single NVIDIA RTX A4000 GPU. 

\subsection{Experimental setup}
To ensure a fair comparison, all policies are constrained to operate under the same input and output specifications. In particular: they all receive raw LiDAR scans as their primary exteroceptive input, and they must return continuous control commands bounded within the triangular feasible action space of the differential-drive robot.

For baselines that natively require the explicit states of surrounding humans (e.g., positions and velocities) rather than raw sensor data, we couple them with the pre-trained perception module of JESSI. This allows for a fair comparison as the same perception is used for all methods that require it. To filter out noise and false positives, only predicted human tracks with a confidence score higher than $0.5$ are passed to the state-based policies. For algorithms requiring a deterministic human radius, a constant human radius of $0.3$\,m is used, matching the simulation parameters. Finally, for purely social policies that lack a native fallback when no humans are detected (i.e., all confidence scores below $0.5$), a heuristic action that drives the robot directly toward the local goal is employed. All the policies are evaluated across two distinct types of experimental settings, varying the number of humans (from 1 to 10) and static obstacles (from 1 to 5) to assess their scalability and robustness.

\begin{figure}
    \centering
    \includegraphics[width=0.99\columnwidth]{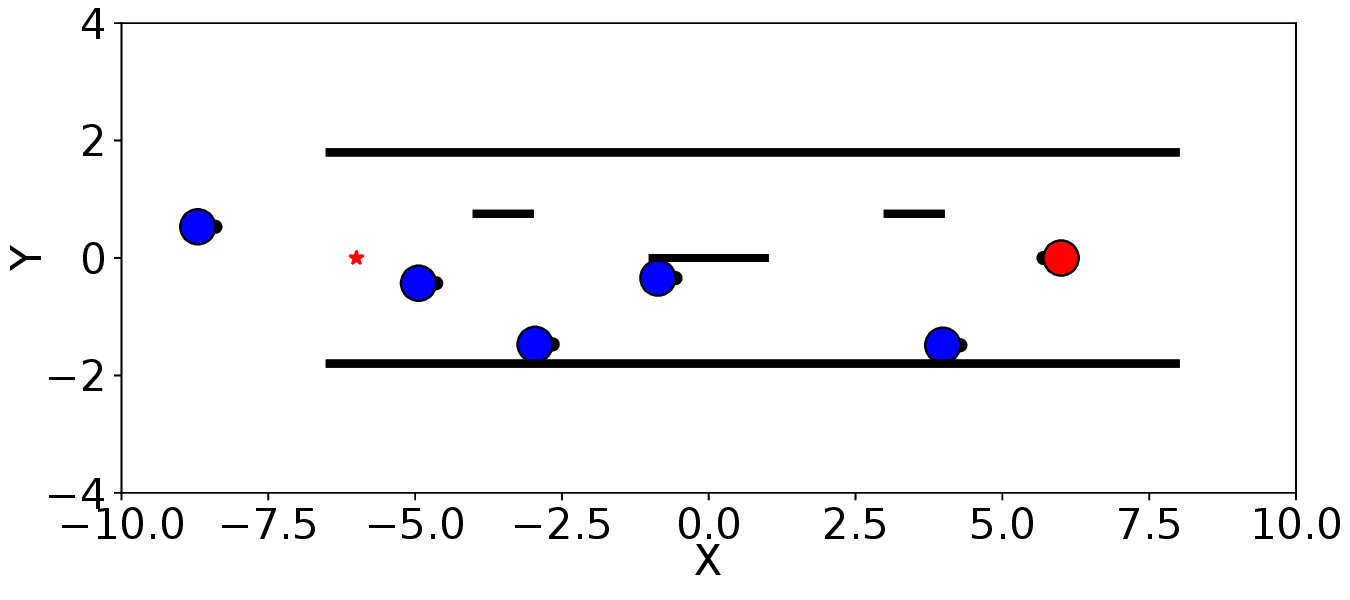}
    \vspace{-6mm}
    \caption{Example of one scenario in the first experimental setup (parallel traffic) with 5 humans (in blue) and 5 box-shaped obstacles (in black). The robot (circle) and its goal (star) are depicted in red. Humans continuously flow in the corridor in the opposite direction of the robot.\vspace{-4mm}}
    \label{fig:first_experiment}
\end{figure}

In the first setting, the policies are evaluated in 6 scenarios already used during training (Circular Crossing, Parallel Traffic, Perpendicular Traffic, Delayed Circular Crossing, Robot Crowding, Crowd Navigation) and 3 unseen scenarios (Corner Traffic, Door Crossing, Crowd Chasing) \cite{francis2023principles}. For an episode to be successful, the robot must reach its goal without collisions before a specified time limit. An example of the parallel traffic scenario is shown in Fig. \ref{fig:first_experiment}.

\begin{figure}
    \centering
    \includegraphics[width=0.7\columnwidth]{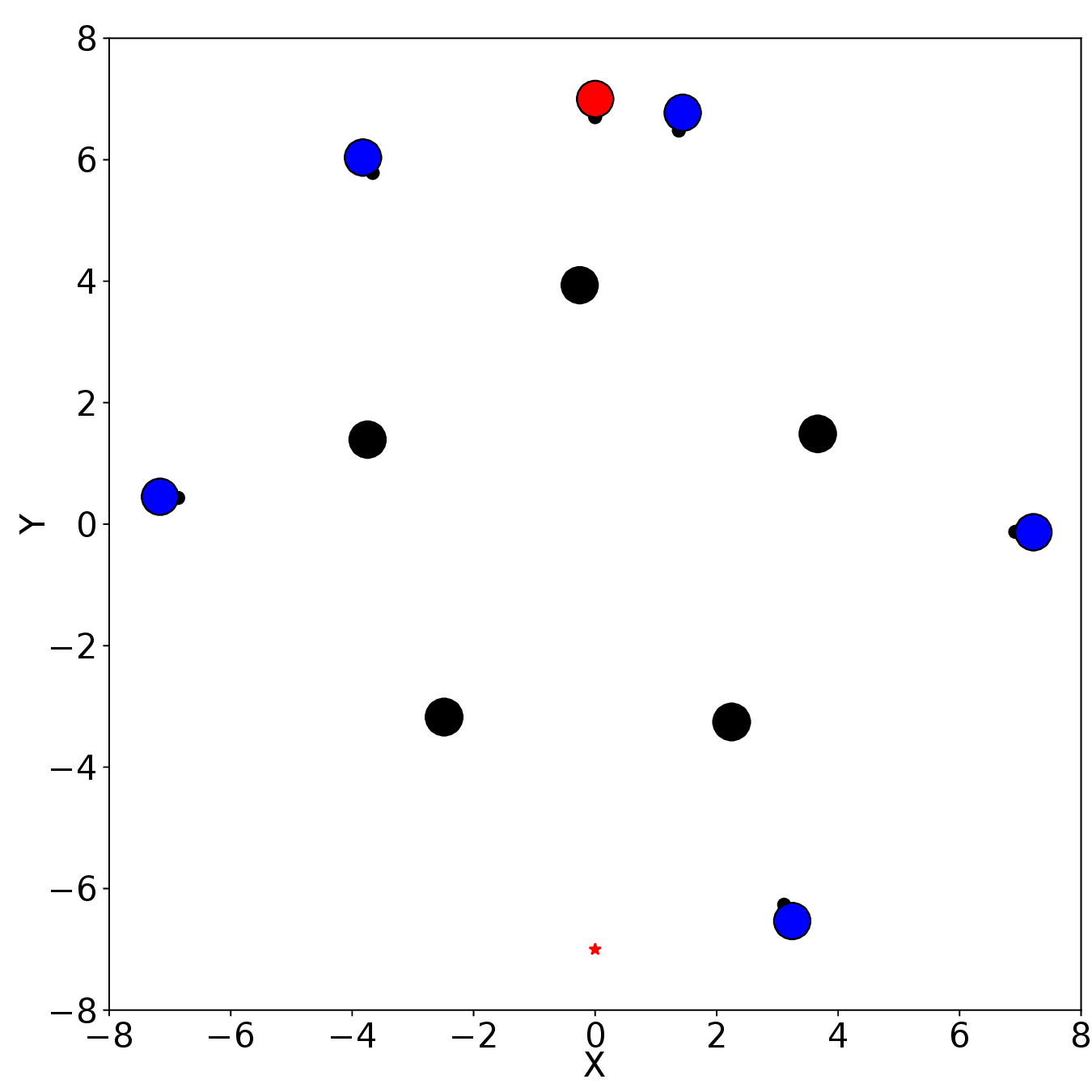}
    \vspace{-3mm}
    \caption{Example of the second experimental setup with 5 humans (in blue) and 5 circular obstacles (in black). The robot (circle) and its goal (star) are depicted in red. Humans continuously flow in the center of the scene.\vspace{-5mm}}
    \label{fig:second_experiment}
\end{figure}

Since certain social navigation baselines cannot natively handle static obstacles, the second experimental setting is designed to include them in our evaluation: a specific scenario where all static obstacles are modeled as stationary humans (with the same radius as moving humans) is employed. This allows purely social policies to navigate the space with both dynamic and static obstacles. In this scenario, wider spaces are available as walls or other cluttering static obstacles are not present, as shown in the example in Fig. \ref{fig:second_experiment}.

\subsection{Evaluated Baselines}
The three proposed trained policies (JESSI-MT, JESSI-MD, JESSI-PO) are compared against a set of model-based and learning-based navigation policies:

\subsubsection{Model-based baselines}
 the Dynamic Window Approach \textbf{DWA} \cite{fox2002dynamic} is a classic, widely adopted local planner that samples velocities in a reachable dynamic window and scores them based on clearance and goal heading (in our experiments, we assume the robot accelerations are not bounded). Model Predictive Path Integral control \textbf{MPPI} \cite{williams2017model} is a state-of-the-art sampling-based controller, frequently used in modern autonomous navigation stacks. A recently proposed Dynamic Risk-Aware extension of the latter (\textbf{DRA-MPPI} \cite{trevisan2025dynamic}), that incorporates a collision risk cost for human interactions, is also considered. While \cite{trevisan2025dynamic} assumes the knowledge of humans' position and velocities and a fixed uncertainty growth for human positions along the prediction horizon, we enhance it by feeding the dynamically estimated means and covariance matrices directly from the JESSI perception module, computing accurate joint probability of collision.

\subsubsection{End-to-end learning baselines}
a \textbf{VANILLA-E2E} baseline (similar to \cite{pfeiffer2017perception}), that processes the LiDAR stack through a multi-layer perceptron, mapping raw data directly to actions. A variant of the previous baseline that incorporates the Dirichlet action space parameterization and the static obstacle bounding algorithm proposed in \cite{van2025safe}, also employed by JESSI, is \textbf{BOUNDED VANILLA-E2E}.

\subsubsection{Humans state-based learning baselines}
Collision Avoidance with Deep Reinforcement Learning \textbf{CADRL} \cite{CADRL} is an early influential social navigation policy that solves an optimization problem online to find the action (over a discrete set) that maximizes the sum of the current reward and the value of the future state, only considering one human (but can be extended to multiple humans through a max-min operator). Socially Aware Reinforcement Learning \textbf{SARL} \cite{SARL} is an evolution of CADRL that utilizes an attention mechanism to pool pairwise human-robot interactions. Like CADRL, it does not explicitly model static obstacles. \textbf{SARL*} \cite{li2019sarl} is an extension of the latter that filters out discrete actions leading to collisions with static obstacles. 
Finally, \textbf{DIR-SAFE} \cite{van2025safe} is a recently proposed RL-based local planner utilizing Dirichlet distributions and safe action-space bounding. All these policies assume perfect knowledge of human states, which, in the following benchmark, are instead supplied by the pre-trained JESSI perception module.

\subsection{Evaluation Metrics}
To quantitatively assess the performance of all tested policies, we employ a comprehensive set of metrics that capture navigation efficiency, safety, and social compliance. Success Rate (SR),  Timeout Rate (TR), Collision with Obstacle Rate (CR-O), Collision with Human Rate (CR-H) indicate the percentage of episodes where the robot successfully reaches all waypoints within the maximum allowed time (SR), fails to do so due to reaching maximum time for completion (TR), fails due to a collision with an obstacle (CR-O) or due to a collision with a human (CR-H).  Time to Goal (TtG) is the average navigation time required to complete successful episodes. Linear (LJ) and Angular Jerk (AJ) are the time derivatives of the linear and angular accelerations. These metrics quantify the smoothness of the generated controls. Smoother trajectories are preferable from both a social and a navigation point of view. Space Compliance (SC) is a social metric evaluating the fraction of the trajectory in which the robot keeps a distance greater than $0.5$~m from all humans.

\subsection{Experimental results}
Table \ref{tab:first_experimental_setup_results} shows the results obtained on the first experimental setup. In terms of success rate, the top three performers are JESSI-PO, JESSI-MD and JESSI-MT in decreasing order. Among these three, JESSI-MT shows the highest space compliance, indicating that it is more capable of maintaining a higher distance from humans throughout the trajectory. This suggests that allowing gradients from the reinforcement learning objective to flow back through the perception backbone alongside the supervised perception loss, slightly improves social performance.
In terms of time to goal and path smoothness the vanilla E2E methods achieve the best performance, indicating that they are the most proficient planners in terms of navigation efficiency. The bounded version of the vanilla E2E policy (BV-E2E) shows a much lower collision with obstacles rate with respect to its standard version (V-E2E), demonstrating the effectiveness of the bounding algorithm (which is also employed in JESSI). Regarding DWA, a very high angular jerk is measured during the experiments, indicating that the policy performs poorly in social terms. Notably, DRA-MPPI achieves the highest space compliance but it also shows one of the highest timeout rates. Empirically, it has been seen that when the uncertainty of humans' position grows too high throughout the prediction horizon, DRA-MPPI fails to reach the goal due to all sampled trajectories having high probability of collision \cite{trautman2010unfreezing}. With regards to SARL* instead, it is observed that the action filtering is not as effective when dealing with noisy inputs (in fact, it shows a high rate of collisions with obstacles), while the same does not happen for DIR-SAFE (at least not at the same scale) as the bounding algorithm is naturally conservative.

\begin{table}[thpb]
\centering
\caption{First experimental setup results}
\resizebox{\columnwidth}{!}{
    \begin{tabular}{ccccccccc}
    \toprule
    Policy & \makecell{SR \\ (\%)} & \makecell{CR-H \\ (\%)} & \makecell{CR-O \\ (\%)} & \makecell{TR \\ (\%)} & \makecell{TtG \\ (s)} & \makecell{LJ \\ (m/s$^3$)} & \makecell{AJ \\ (rad/s$^3$)} & \makecell{SC \\ (\%)} \\
    \midrule
    JESSI-MT & 94.2 & 0.0 & 1.8 & 4.0 & 18.25 & 1.30 & 4.84 & 90.1 \\
    JESSI-MD & 94.9 & 0.3 & 1.7 & 3.1 & 17.88 & 1.46 & 4.88 & 88.2 \\
    JESSI-PO & 95.1 & 0.1 & 2.2 & 2.6 & 18.03 & 1.77 & 5.45 & 87.9 \\
    V-E2E & 86.3 & 0.2 & 12.6 & 0.9 & 16.93 & 0.76 & 2.27 & 87.6 \\
    BV-E2E & 92.9 & 0.2 & 1.5 & 5.4 & 17.01 & 1.04 & 1.91 & 85.0 \\
    DIR-SAFE & 73.2 & 1.1 & 4.5 & 21.2 & 23.17 & 2.99 & 10.82 & 92.1 \\
    SARL* & 63.5 & 0.2 & 23.2 & 13.1 & 19.39 & 3.45 & 15.19 & 92.3 \\
    DWA & 67.9 & 0.0 & 0.0 & 32.1 & 17.83 & 2.35 & 43.05 & 79.7 \\
    MPPI & 73.0 & 0.1 & 0.4 & 26.5 & 22.76 & 5.89 & 25.67 & 87.7 \\
    DRA-MPPI & 79.6 & 0.1 & 1.8 & 18.5 & 22.13 & 3.53 & 9.58 & 93.5 \\
    \bottomrule
    \end{tabular}
}
\label{tab:first_experimental_setup_results}
\end{table}\vspace{-2mm}

In Table \ref{tab:second_experimental_setup_results} the results of the second experimental setup are shown. Also in this setup, the three JESSI policies achieve the highest success rate with JESSI-MT having the highest space compliance among them. Again, the idea that providing an intermediate representation of the state of humans rather than implementing a naive E2E approach seems to be beneficial in terms of social performance. In this specific scenario, it might be more evident (especially with respect to BV-E2E) as wider spaces allow for a more social behavior, while highly cluttered environments forces the robot to be more navigation-oriented. The performance obtained by JESSI policies benefits from their ability to leverage the full probabilistic distribution of the perception output. By accounting for the inherent uncertainty of human state estimates (rather than relying solely on mean predictions) the policy can navigate more cautiously when detections are noisy. Concerning SARL and CADRL, it is observed that the high rate of collision with obstacles (modeled as static humans in this setup) are due to the fact that sometimes the pre-trained perception of JESSI does not detect static humans as such. Thus, the downstream policies that only consider humans for planning will not be able to perceive them. Instead, JESSI also relies on processed LiDAR features for planning, meaning that even if the perception module fails, the policy will be able to perceive such obstacles through other inputs. Regarding DIR-SAFE, SARL and DRA-MPPI, a high space compliance is observed but it comes at a cost in terms of success rate (especially for SARL).

\begin{table}[thpb]
\centering
\caption{Second experimental setup results}
\resizebox{\columnwidth}{!}{
    \begin{tabular}{ccccccccc}
    \toprule
    Policy & \makecell{SR \\ (\%)} & \makecell{CR-H \\ (\%)} & \makecell{CR-O \\ (\%)} & \makecell{TR \\ (\%)} & \makecell{TtG \\ (s)} & \makecell{LJ \\ (m/s$^3$)} & \makecell{AJ \\ (rad/s$^3$)} & \makecell{SC \\ (\%)} \\
    \midrule
    JESSI-MT & 99.9 & 0.0 & 0.0 & 0.1 & 20.55 & 0.97 & 2.96 & 89.7 \\
    JESSI-MD & 99.8 & 0.0 & 0.0 & 0.2 & 19.82 & 1.11 & 2.98 & 87.5 \\
    JESSI-PO & 100.0 & 0.0 & 0.0 & 0.0 & 19.73 & 1.25 & 3.07 & 87.9 \\
    V-E2E & 99.0 & 0.0 & 1.0 & 0.0 & 18.82 & 0.65 & 1.61 & 87.3 \\
    BV-E2E & 99.8 & 0.0 & 0.2 & 0.0 & 17.58 & 0.83 & 1.31 & 80.7 \\
    DIR-SAFE & 98.1 & 0.0 & 0.1 & 1.8 & 22.65 & 2.63 & 7.52 & 91.4 \\
    SARL* & 92.8 & 0.0 & 5.5 & 1.7 & 18.01 & 3.23 & 9.13 & 90.8 \\
    SARL & 80.0 & 0.0 & 20.0 & 0.0 & 17.43 & 3.13 & 8.22 & 91.4 \\
    CADRL & 63.5 & 0.0 & 36.3 & 0.2 & 19.34 & 3.26 & 7.77 & 81.6 \\
    DWA & 47.3 & 0.0 & 0.0 & 52.7 & 20.10 & 1.82 & 42.19 & 76.3 \\
    MPPI & 99.7 & 0.0 & 0.3 & 0.0 & 23.84 & 5.57 & 22.91 & 82.4 \\
    DRA-MPPI & 96.7 & 0.0 & 0.0 & 3.3 & 24.94 & 3.52 & 9.53 & 95.2 \\
    \bottomrule
    \end{tabular}
}
\label{tab:second_experimental_setup_results}
\end{table}

\section{REAL WORLD EXPERIMENTS}\label{sec:real_world}
\begin{figure}[t]
    \centering
    \includegraphics[width=.78\columnwidth]{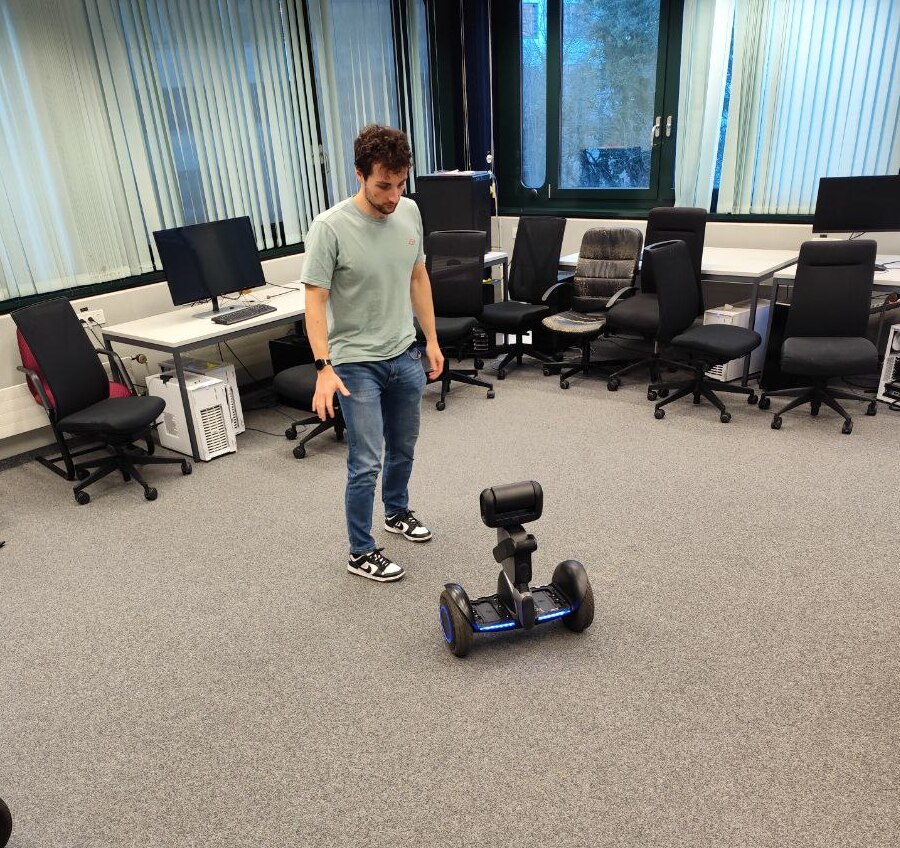} 
    \caption{Experimental setup of the real world test with the Loomo Segway robotic platform.\vspace{-8mm}}
    \label{fig:real_test}
\end{figure}

The trained policy JESSI-MT has been fine-tuned and tested and on a Turtlebot4 and on a Loomo Segway robotic platform. The Turtlebot4 is equipped with a 360° LiDAR with a maximum range of 12 meters. The deployment on the Loomo has been specifically designed to test the sensor-agnostic capabilities of the JESSI architecture (recall that trained policies are not tied to any specific LiDAR configuration). Since the Loomo is not equipped with a LiDAR, the distance measurements have been taken from the onboard depth camera, after converting the depth image into a 2D LiDAR scan with a field-of-view of approximately 60° and a maximum range of 4 meters. The experimental setup using the Loomo is shown in Fig. \ref{fig:real_test}. To optimize the performance on the specific LiDAR setting for each robot, a fine-tune RL training in simulation has been performed. A ROS2 interface has been used to run inferences on JESSI on a remote device. The average inference time has been measured as $20$~ms on an Intel(R) Core(TM) i5-7200U CPU @ 2.50GHz. Such a low latency is achieved thanks to the JAX parallelization and JIT compilation. The entire codebase for training and testing, the trained policies and some videos demonstrating JESSI capabilities in simulation and the real world are available at \url{https://github.com/TommasoVandermeer/social-jym/tree/jessi}.\vspace{-3mm}

\section{CONCLUSIONS}\label{sec:conclusions}
In this work, we presented JESSI, a novel JAX-based end-to-end reinforcement learning framework for safe and socially compliant robot navigation. By directly mapping raw LiDAR scans to kinematically feasible continuous control commands through a sensor-agnostic architecture, JESSI bridges the gap between idealized state-based planners and practical sensor-based autonomy.
The extensive simulated evaluations across diverse human-populated scenarios demonstrate the effectiveness of our approach against several state-of-the-art baselines. A key finding of the experiments is that integrating a supervised perception signal seems to enhance the social performance of the policy. Unlike vanilla end-to-end methods that treat humans merely as generic geometric obstacles, JESSI utilizes an intermediate probabilistic representation of human states, which inherently grounds the RL policy in a physically meaningful latent space. As shown in the numerical results, providing this explicit semantic understanding of human presence directly translates to higher success rate and slightly higher space compliance. 
The numerical evaluation highlights the benefit of modeling the continuous action space through Dirichlet distributions coupled with a deterministic bounding algorithm. This architectural choice reduces collision rates with static obstacles, as evidenced by the performance gap between the bounded and unbounded baseline policies.
The successful deployment of the trained policy on a Turtlebot4 with a 360° LiDAR and Loomo Segway platform utilizing a depth camera converted into a 2D LiDAR scan, confirms the real-world applicability and computational efficiency of the JESSI architecture.




\bibliographystyle{ieeetr}
\bibliography{refs}

\end{document}